\documentclass[11pt,a4paper]{article}
\usepackage[hyperref]{emnlp-ijcnlp-2019}
\usepackage{times}
\usepackage{latexsym}
\usepackage{graphicx}
\usepackage{url}
\usepackage{xcolor}
\usepackage{amsfonts}
\usepackage{amsmath}
\usepackage{booktabs}
\usepackage{lipsum}
\usepackage{footmisc}
\usepackage{float}
\usepackage{stfloats}

\aclfinalcopy 

\title{Point-Or-Generate Dialogue State Tracker}

\author{Song Xiaohui \\
  Institute of Information Engineering \\
  Chinese academy of Sciences\\
  Beijing, China \\
  {\tt songxiaohui@iie.ac.cn} \\\And
  Hu Songlin \\
  Institute of Information Engineering \\
  Chinese academy of Sciences\\
  Beijing, China \\
  {\tt husonglin@iie.ac.cn} \\}

\date{}
\begin{document}
\maketitle
\begin{abstract}
  Dialogue state tracking is a key part of a task-oriented dialogue system, which estimates the user's goal at each turn of the dialogue. In this paper, we propose the Point-Or-Generate Dialogue State Tracker (POGD). POGD solves the dialogue state tracking task in two perspectives: 1) point out explicitly expressed slot values from the user's utterance, and 2) generate implicitly expressed ones based on slot-specific contexts. It also shares parameters across all slots, which achieves knowledge sharing and gains scalability to large-scale across-domain dialogues. Moreover, the training process of its submodules is formulated as a multi-task learning procedure to further promote its capability of generalization. Experiments show that POGD not only obtains state-of-the-art results on both WoZ 2.0 and MultiWoZ 2.0 datasets but also has good generalization on unseen values\footnote{Values that didn't appear in the training data called unseen values.} and new slots.

\end{abstract}

\section{Introduction}
\label{intro}
Task-oriented dialogue system interacts with users in natural language to accomplish tasks such as restaurant reservation or flight booking. The goal of dialogue state tracking is to provide a compact representation of the conversation at each dialogue turn, called \textit{dialogue state}, for the system to decide the next action to take. A typical dialogue state is consist of goal of user, action of the current user utterance  (\texttt{inform}, \texttt{request} etc.) and dialogue history \cite{young2013pomdp}. User's goal is the most important among them, it is a set of slot-value pairs. All of these are defined in a particularly designed \textit{ontology} that restricts which slots the system can handle, and the range of values each slot can take. The \textit{joint goal} is a set that accumulates turn goals up to the current turn. Accuracy of the joint goal is a rigorous evaluation, because each error slot-value pair may propagate to the next few turns. In this paper, we focus on tracking the user's goal and use the joint goal accuracy to evaluate our model.

\begin{figure}[t]
\centering
\includegraphics[width=0.45\textwidth]{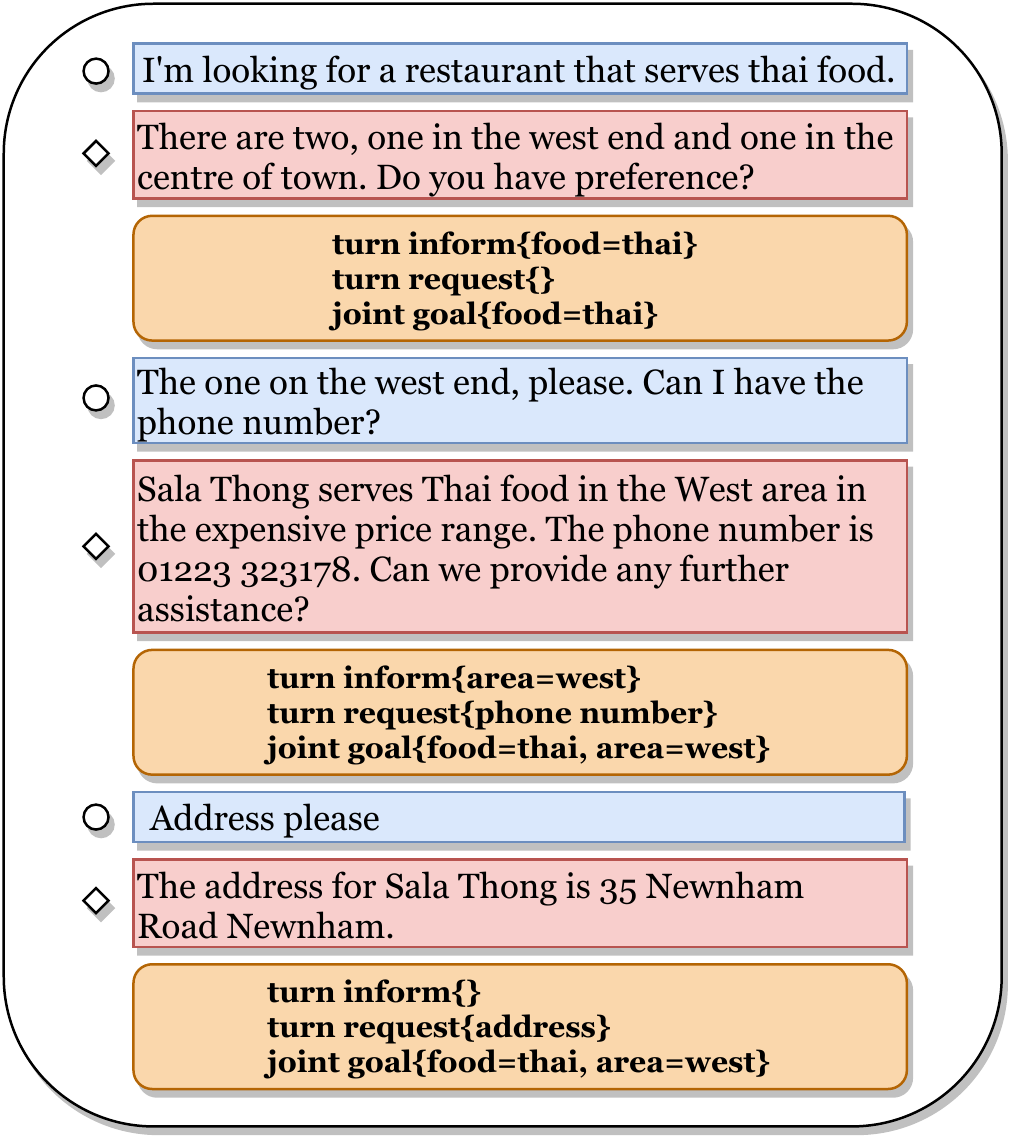}
\caption{An example from the WoZ restaurant reservation corpus. 
Each dialogue turn contains a user utterance (labeled with circle)
and a system utterance (labeled with rhombic), we annotated 
\textit{turn inform},
\textit{turn request} and \textit{joint goal} in rounded rectangles, respectively.}
\label{figure1}
\end{figure}

Considering an example of restaurant reservation, users can \textit{inform} the system some restrictions of their goals (\textit{e.g.}, \texttt{inform(food = thai)}) or \textit{request} further information they want (\textit{e.g.}, \texttt{request(phone number)}) at each turn. Figure~\ref{figure1} shows an annotated dialogue in WoZ 2.0 dataset.

 \begin{figure*}[hbtp]
  \centering
  \includegraphics[width=\textwidth]{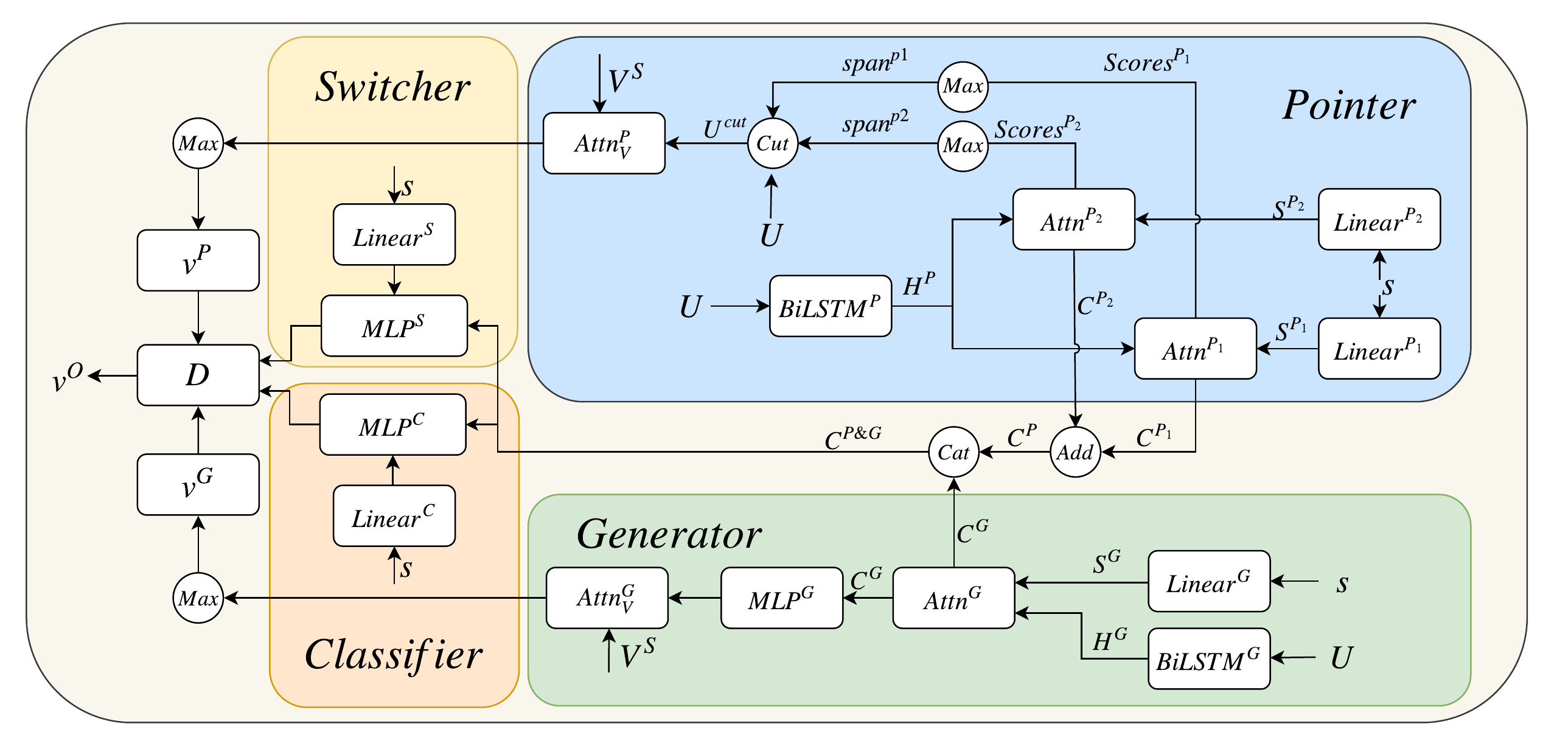}
  \caption{The architecture of POGD. The four modules (\textit{i.e.} Pointer, Generator, Switcher and Classifier) are jointly trained via multi-task learning. Inputs are user and system utterance $U$, considering slot $s$, and its value set $V^{S}$. Losses are calculated based on $Scores^{P_1}$, $Scores^{P_2}$, the final predicted value $\mathit{v}^{O}$, and the outputs of Switcher and Classifier.}
  \label{figure2}
  \end{figure*}

  In many task-oriented dialogue system applications we have built in the industry, the state tracker faces several challenges. First, with the complexity of tasks increases, more slots and values can be used during the conversation, the state tracker needs to be more \textbf{accurate} otherwise the system may fail to accomplish tasks. Second, the number of slots or tasks in the ontology may be large, the state tracker requires \textbf{scalability} to large-scale multi-domain dialogues. Third, when new values or slots are introduced to the system, it is hard to collect lots of labeled data, so the ability of \textbf{few-shot} or even zero-shot learning on newly added data is required. To tackle these challenges, we propose a scalable multi-domain dialogue state tracker, \textbf{P}oint-\textbf{O}r-\textbf{G}enerate \textbf{D}ialogue State Tracker (POGD), which could handle unseen values and easy to generalize to new slots.

  Let's discuss the designation of our proposed POGD model. By taking a deep look at the dialogue data examples, we find that the user's goal can be split into explicit and implicit cases\footnote{In this paper, we think a value is explicitly expressed if it has a Levinstein distance less than 3 with it in the ontology.}. The explicit case means that the value is expressed the same way as it in the ontology, otherwise, it is the implicit case. The explicit case usually needs information searching and matching, while implicit one requires generation. It seems natural to introduce Pointer-Generator Network \cite{see2017get} to improve the accuracy. However, the distribution of the above two cases is unbalanced, \textit{i.e.} matching is used much more than generating. Furthermore, values in the implicit case are also in an imbalanced distribution\footnote{In WoZ 2.0 there are 14.65\% implicit values, and in MultiWoZ 2.0 the number is 10.50\%. Among implicit values, the value \texttt{don't care} accounts 82.49\% in WoZ 2.0 and 16.64\% in MultiWoZ 2.0, respectively.}. Even in the explicit case, values may not be expressed exactly the same as it in the ontology, we called it rephrasing\footnote{In the explicit case, rephrased values take a proportion of 29.32\% in WoZ 2.0 and 32.1\% in MultiWoZ 2.0, respectively.} (\textit{e.g.}, \texttt{moderately} in user's utterance and \texttt{moderate} in the ontology). For the above reasons, straightforward utilization of PGNet might lead to chaos. 

  We attempt to think the same way as PGNet---point out slot values from the user's utterance or generate them based on slot-specific contexts. Figure~\ref{figure2} shows the architecture of POGD. POGD is designed as follows: A \textbf{Pointer} is trained to figure out values explicitly mentioned in utterances, while a \textbf{Generator} is trained to infer implicitly expressed ones. Different from PGNet, the Pointer and Generator are decomposed into two separate modules, and a \textbf{Switcher}---a binary classifier is adapted to enable adaptive switching between them to fit the distribution. To handle rephrasing values in the explicit case and imbalanced distribution of implicit values, a looking up in the value set is performed via attention mechanism in both Pointer and Generator. Finally, a \textbf{Classifier} is applied to determine whether a slot is relevant to an utterance or not. POGD shares all parameters across all slots to be scalable, and to achieve knowledge sharing. The training of the four modules is formulated as a multi-task learning procedure to promote the capability of generalization. POGD achieves a state-of-the-art performance of joint goal accuracy on the small-scale WoZ 2.0 \cite{wen2016network} dataset. On the large-scale multi-domain dataset MultiWoZ 2.0 \cite{budzianowski2018multiwoz}, our model obtains 39.15\% joint goal accuracy, outperforming prior work by 3.57\%. The \textbf{contributions} of this paper are three folds:
\begin{itemize}
    \item We address the DST problem by solving two easier subtasks, 1) point out explicitly expressed slot values in user's utterance and 2) generate implicitly expressed ones based on slot-specific contexts.
    \item We propose a scalable multi-domain dialogue state tracker POGD, which obtains state-of-the-art results.
    \item To our knowledge, this is the first attempt to evaluate the capability of generalization of a dialogue state tracker on unseen slot values and new slots.
  \end{itemize}

  The rest of this paper is structured as follows: In Section~\ref{relatedwork} we give a brief review of previous works in the field and show their limitations. The architecture and other details of our model are described in Section~\ref{method}. We perform several experiments to show advantages of POGD in Section~\ref{experiments}. Followed by conclusions in Section~\ref{conclusion}.

\section{Related Work}
\label{relatedwork}
Traditional works on DST such as rule-based models \cite{wang2013simple,sun2014generalized}, statistical models \cite{thomson2010bayesian, young2010hidden, lee2013recipe,lee2013structured,sun2014sjtu,xie2015recurrent} failed to produce satisfactory performance and now rarely discussed. Approaches using a separate Spoken Language Understanding (SLU) module \cite{henderson2012discriminative,zilka2015incremental,mrkvsic2015multi} suffered from error accumulation from the SLU, and relied on hand-crafted semantic dictionaries or delexicalization---replace the values in the utterance with general tags to achieve generalization. 

Recently, deep-learning showed its power to the dialogue state tracking challenges (DSTCs) \cite{williams2013dialog,henderson2014second,henderson2014third}. A variety of DL-based methods were proposed: Neural Belief Tracker (NBT) \cite{mrkvsic2016neural} applied representation learning to learn features as opposed to hand-crafting features; PtrNet \cite{xu2018end} aimed to handle unknown values; GLAD \cite{zhong2018global} and its further improved version GCE \cite{nouri2018toward} addressed the problem that extraction of rare slot-value pairs; another improvement of GLAD \cite{sharma2019improving} used the relevant past user's utterance to get a better performance. But parameters of these models increased with the number of slots. For these scalable approaches, \citet{ramadan2018large} tried to share parameters across slots, but their model needs to iterate all slots and values in the ontology at each dialogue turn. \citet{rastogi2017scalable} generated a fix set of candidate values using a separate SLU module, so suffered from error accumulation. StateNet \cite{ren2018towards} reduced the dependence of ontology but no verification was done in their work. 

\section{Method} 
\label{method}
In this section, we discuss the details of our model. We formulate state tracking in the way that predicts the turn state like GLAD \cite{zhong2018global}, but we do not use the previous system acts in POGD's inputs. At each dialogue turn, POGD's input is consist of utterance $U$, a slot under considering $s$ and its value set $V^{S}$. For utterance $U$, we simply concatenate user and system utterance ($U^{usr}$ and $U^{sys}$) by a particular symbol ${<}$\texttt{usr}${>}$.
\begin{equation}
  U=U^{sys} \oplus {<}\texttt{usr}{>} \oplus U^{usr}
\end{equation}

\subsection{Pointer}
The Pointer is inspired by previous work PtrNet \cite{xu2018end}, which addressed the DST problem using Pointer Networks \cite{vinyals2015pointer}. We use a bidirectional LSTM \cite{hochreiter1997long} to get the encoding $H^{P}$ of $U$ the same way as PtrNet.
\begin{equation}
H^{P} = BiLSIM^{P}(U)\in \mathbb{R}^{n\times d_h}
\end{equation}
Where $n$ is number of words in $U$ and $d_h$ is the dimension of LSTM hidden states. Different from PtrNet, Pointer module is not designed in an encoder-decoder structure, we use two Linears ($\mathrm{Linear}(X) = WX+b $) ${Linear^{P_1}}$ and $Linear^{P_2}$  to encode slot information.
 \begin{equation}
  S^{P_i} = Linear^{P_i}(s)\in \mathbb{R}^{1 \times 2d_h} \quad i\in \{1,2\}
 \end{equation}
 Then, the starting position $span^{p_1}$ and the ending position $span^{p_2}$ of predicted value are computed via $Attn^{P_i}$ in Figure~\ref{figure2} as following, 
\begin{align} 
  \alpha^{P_{i}}_{j} &= v^{\mathrm{T}}\mathrm{tanh}(\mathrm{Linear}(H^{P}_{j} + S^{P_i}))\\
  Scores^{P_{i}}_{j}  &= \exp{\alpha^{P_{i}}_{j}}/\sum_{j=1}^{n}\exp{\alpha^{P_{i}}_{j}}\\
  span^{p_i} &= \mathop{\arg\max}\limits_{j} Scores^{P_{i}}_{j}
\end{align}
where $1\le i \le 2, 1\le j \le n$ and we define a trainable parameter $v$ and a $\mathrm{Linear}$ in each attention module. Note that there is no guarantee that $span^{p_1} \le span^{p_2}$, when $span^{p_1} > span^{p_2}$, it indicates that Pointer's output is not reliable, we will set Switcher's output to choose Generator's output as the final predicted value.
 
We get a $U^{cut}$ by summing the word embeddings between $span^{p_1}$ and $span^{p_2}$ in $U$ and compute its unit vector. It is used to calculate cosine similarity with embeddings of values, as shown in Figure~\ref{figure2}. We find that values and their rephrased have similar embeddings, which is why the Pointer can overcome rephrasing values via attention mechanism. 
\begin{align}
  U^{cut}_{origin} &= \sum_{j=span^{p_1}}^{span^{p_2}}U_j\\
  U^{cut} &= U^{cut}_{origin} / \left\|U^{cut}_{origin}\right\|_2
\end{align}
The cosine similarity between $U^{cut}$ and values in value set $V^{S}$ is calculate using dot attention \cite{luong2015effective}, 
\begin{gather}
  v^{P} = \mathop{\arg\max}\limits_{V^{S}_{j}} U^{cut} \cdot V^{S}_{j}
\end{gather}
where $j$ is in range of the number of values in $V^{S}$, we select the value with max attention score $v^{P}$ as the output of Pointer. We also calculate a context $C^{P}$ by following steps for further using in Classifier and Switcher.
 \begin{align}
   C^{P_{i}} &= \sum_{j=1}^{n} Scores_{j}^{P_{i}} H_{j}^{P}\\
   C^{P} &= C^{P_{1}} + C^{P_{2}}
 \end{align}

\subsection{Generator}
The Generator works similar to Pointer, as described in Figure~\ref{figure2},
 after encoding utterance and slot and computing attention scores the same way as Pointer, a slot-specific context $C^{G}$ is computed as follows.
 \begin{align}
  Scores^{G} &= Attn^{G}(S^{G}, H^{G})\\
  C^{G} &= \sum_{j=1}^{n} Scores_{j}^{G}H_{j}^{G}
 \end{align}
Different from Pointer, Generator is designed to infer the implicit values
based on slot-specific contexts, so it needs to handle more complex similarity than the word embeddings similarity computed in $Attn^{P}_{V}$ in Pointer. $C^{G}$ is transformed by a multilayer perceptron (\textrm{MLP}) \cite{gardner1998artificial} with ReLU \cite{nair2010rectified} activation, let $\mathrm{MLP^{i}_{ReLU}}$ denote a $i$-layers MLP with ReLU activation at each hidden layer, transformed context is computed as following.
\begin{gather}
  C^{G^{\prime}}  = \mathrm{MLP^{2}_{ReLU}}(C^{G})
\end{gather}
Attention $Attn_{V}^{G}$ performed in Generator is to deal with the imbalance of implicit values. Traditional sequence to sequence models like PGNet \cite{see2017get} used a Linear to select final output from vocabulary, which may suffer from imbalanced distribution of values, because each value in vocabulary corresponding to a weight in the Linear. So instead of using a Linear, we use attention mechanism, the trainable parameters $v$ and $Linear$ in $Attn_{V}^{G}$ are shared across all values, which reduces the impact of data imbalance on performance. $Attn_{V}^{G}$ does't perform dot attention as $Attn^{P}_{V}$ in Pointer but perform the same as $Attn^{G}$. We also choose the value that has the max score $v^{G}$ as the final output of Generator.
\begin{align}
   Scores_{V}^{G} &= Attn_{V}^{G}(C^{G^{\prime}}, V^{S})\\
   v^{G} &= \mathop{\arg\max}\limits_{V_{j} ^{S}} Scores_{V}^{G}
\end{align}

\subsection{Classifier \& Switcher}
Classifier and Switcher are two binary classifiers that have the same 
architecture---a 3-layers perceptron with dropout \cite{srivastava2014dropout}
 at the hidden layer, and they also have the same inputs except for the encoding of slots. But the goal of the two modules are different: Switcher is trained to choose the final predicted value from Pointer's output $v^{P}$ and Generator's output  $v^{G}$, while Classifier is trained to determine whether the slot under considering $s$ is relevant to the input utterance $U$ or not. Let's discuss how Switcher works, and the Classifier acts the same.

 The Switcher's input is consist of three parts: Pointer's attention context $C^{P}$, Generator's attention context $C^{G}$, and the encoding of slot. As shown in Figure~\ref{figure2}, they are simply concatenated together. let $I_{Switcher}$ denote the input of Switcher, the Switcher works as following.
 \begin{align}
   I_{Switcher}&=C^{G} \oplus C^{P} \oplus Linear^{S}(s) \\
   h &= \mathrm{MLP^{2}_{ReLU}}(I_{Switcher})\\
   h^{\prime} &= \mathrm{dropout}(h)\\
   output^{S} &= \mathrm{sigmoid}(\mathrm{Linear}(h^{\prime}))
 \end{align}
 where $Linear^{S}$ is the slot encoder of Switcher as shown in Figure~\ref{figure2}. The Switcher let Pointer and Generator learn simplified tasks---Pointer learns to point only on pointable values and Generator learns to generate only on implicit values.
 
  The final output produced by the $\mathrm{sigmoid}$ \cite{lecun2012efficient} function. In our experiments, the final predicted value $v^{O}$ is produced as follows.
\begin{align}
  v^{O\prime} &= v^{P} output^{S} + v^{G}(1-output^{S})\\
  v^{O} &= v^{O\prime} output^{C}
\end{align}
Here $output^{S}$ and $output^{C} \in \{0,1\}$ are the output of Switcher
and Classifier. 

\subsection{Multi-Task Learning}
The four modules in our model are jointly trained via multi-task learning.
This section we discuss how we separate the tasks and which parameters are shared.

We use 5 loss functions to train our model. In Pointer, we use 
$Scores^{p_{i}}$, $i\in \{1,2\}$ as outputs and ground truth starting and ending positions of values as labels to compute two cross entropy losses $Loss^{P_{1}}$ and $Loss^{P_{2}}$. For Switcher and Classifier, they use expected outputs generated from the dataset as labels and perform binary cross entropy losses $Loss^{S}$ and $Loss^{C}$. During the training process, we use Switcher's labels to produce final output $v^{O}$, then a cross entropy loss $Loss^{final}$ is computed between $v^{O}$ the ground truth output of the tracker. Generator is a little particular, it doesn't have its own loss function, as we described in the last subsection, $v^{O}$ only contains Pointer's output $v^{P}$ or Generator's output $v^{G}$, so the error back propagation of $Loss^{final}$ will only go through either Pointer or Generator. We use cut values in Pointer and the indexing data doesn't have gradient so the Pointer needs its own losses.

The Classifier needs negative sampling, which brings more difficulty to the joint training process. Negative samples are randomly selected from unrelated slots with some ratio, we should make them only have effect on the Classifier. When a negative sample go through the model, Switcher's label will set to its output before the back propagation, and other modules' labels will set to zero and ignored from cross entropy loss functions. 

Classifier and Switcher use Pointer and Generator's attention contexts as a part of the input, \citet{caruana1997multitask} showed that multi-task learning can improve generalization, our experiments further verified this argument.

\section{Experiments}

In this section, we perform several experiments to show the advantages of our model. We show joint goal accuracy of POGD on two datasets, and compare to several previous approaches. Then we examine our model's generalization on unseen values, and the capability of few-shot learning on new slots.

\label{experiments}

\subsection{Dataset}
\begin{table}[htbp]
  \centering
  \begin{tabular}{l|l|l}
    \hline
   \textbf{Metric} & \textbf{WoZ} & \textbf{MultiWoZ} \\ \hline
   \textrm{dialogues} & 600 & 8,438   \\
   \textrm{total turns} &4,472  & 115,424   \\ 
   \textrm{average tokens per turn} & 11.24 & 13.18 \\ 
   \textrm{inform slots} &3 &  35  \\ 
   \textrm{tot values} & 99  & 4510  \\ 
   \textrm{unique tokens}& 2,142 & 24,071 \\\hline
  \end{tabular}
  \caption{Details compare between WoZ 2.0 and MultiWoZ 2.0 dataset.}
  \label{datasets}
\end{table}

We mainly use two datasets to evaluate our model, one is a small-scale dataset, the second version of Wizard-of-Oz (\textbf{WoZ 2.0}) \cite{wen2016network} which user's goal is to find a suitable restaurant around Cambridge. Another one is a large-scale multi-domain dataset, the second version \textbf{MultiWoZ 2.0} \cite{budzianowski2018multiwoz}, which contains 6 domains but we only use 5 domains occurred in its test set as \citet{ramadan2018large} did. Both of them are human-machine conversations. A comparison of two datasets showed in Table~\ref{datasets}.

\begin{table*}[!tp]
  \centering
  \begin{tabular}{l|l|l}
  \hline
  DST models & \begin{tabular}[c]{@{}l@{}}Joint Acc.\\ WoZ 2.0 (\%)\end{tabular} & \begin{tabular}[c]{@{}l@{}}Joint Acc.\\ MultiWoZ 2.0 (\%)\end{tabular} \\ \hline
  Belief Tracking: CNN\cite{ramadan2018large} & 85.5 & 25.83 \\ 
  Neural Belief Tracker: NBT-DNN\cite{mrkvsic2016neural} & 84.4 & / \\ 
  GLAD\cite{zhong2018global} & 88.1 & 35.57 \\ 
  GCE\cite{nouri2018toward} & 88.5 & 35.58 \\ 
  GLAD + RC + FS\cite{sharma2019improving} & \textbf{89.2} & / \\ 
  SateNet\cite{ren2018towards} & 88.9 & / \\ \hline
  \textbf{POGD (ours)} & 88.7 & \textbf{39.15} \\ \hline
  \end{tabular}
  \caption{Joint goal accuracy on WoZ 2.0 and MultiWoZ 2.0 test set vs.
  various approaches as reported in the literature.}
  \label{tab:result}
\end{table*}

\subsection{Implementation Details}
We use \textit{semantically specialised} Paragram-SL999 vectors \cite{wieting2015paraphrase} the same as Neural Belief Tracker \cite{mrkvsic2016neural} and do not fine-tune the embeddings of utterance and values.
Slots' embeddings are random initialized. Model is trained using ADAM \cite{kingma2014adam} with learning rate 1e-3. We apply dropout \cite{srivastava2014dropout} with rate 0.3 at word embeddings and hidden layers of $\mathrm{MLPs}$ in Classifier and Switcher. The position labels used to train Pointer are generated the same way as PtrNet \cite{xu2018end}, we use the last occurrence of the reference value in the dialogue history, if a subsequence of an utterance in history has a Levenshtein distance less than 3 with the value, it will be treated as a successful match. Switcher's labels are generated the same time---matching is successful or not. Labels of Classifier need negative sampling, we randomly choose unrelated slots as negative samples, with a probability of 13/30 in MultiWoZ 2.0, and in the WoZ 2.0 dataset we use all unrelated slots as negative samples. We train the model 400 epochs with L2 regularization 2e-7 on WoZ 2.0 and 50 epochs with L2 1e-7 on MultiWoZ 2.0. Other details of data generation and parameters of POGD are described in the Appendix.

\subsection{Performance}
As described in Section~\ref{intro}, the joint goal accuracy is a widely used metric in DST task. Table~\ref{tab:result} shows the joint goal accuracy of our model compares against various previous reported baselines on WoZ 2.0 and MultiWoZ 2.0 test set. POGD achieves competitive result 88.70\% on the small dataset and gets a state-of-the-art result 39.15\% on the large scale dataset.

\subsection{Generalization}
In this section, we will show the POGD model is generalizable in two aspects: unseen values and new slots. New values or called unseen values means they don't appear during the training process. 

\subsubsection{Unseen values}

The generalization of unseen values is evaluated at WoZ 2.0 dataset. We randomly choose values at rate 15\% to 55\% of slot \textit{food} which contains 76 values in total, and delete data contains these values from the train set. We report Precision, Recall and F1 score of these values in the test set, as shown in Table~\ref{tab:unseen}. 

Experiments here are not intend to get the best performance but to prove our POGD model \textbf{can} handle unseen values, so we only train 20 epochs to get results in Table~\ref{tab:unseen}. It is important to emphasize that our POGD can generalize to unseen values without using SLU module or delexicalization like previous works\cite{rastogi2017scalable, ren2018towards}.

\begin{table}[htbp]
  \centering
  \begin{tabular}{l|l|l|l}
  \hline
  unseen(\%) & Precision & Recall & F1 score \\ \hline
  15 (4.15) & 0.9592 & 0.9216 & 0.9400 \\ 
  25 (10.55) & 0.9746 & 0.8647 & 0.9163 \\ 
  35 (11.35) & 0.9565 & 0.8333 & 0.8907 \\ 
  45 (16.31) & 0.9776 & 0.8137 & 0.8881 \\
  55 (24.80) & 0.9132 & 0.8155 & 0.8616 \\ \hline
  \end{tabular}
  \caption{POGD's performance on unseen values. Numbers in brackets indicate the proportion of deleted training examples. Unseen values are randomly selected, and each value takes different proportions in the training data.}
  \label{tab:unseen}
\end{table}

\subsubsection{New slots}

The generalization of new slots is rarely discussed. Recent approaches like NBT \cite{mrkvsic2016neural} and PtrNet \cite{xu2018end} trained model separately for each slot type, when facing a new slot, they need to train an entirely new model. As for those scalable approaches \cite{zhong2018global,ren2018towards,ramadan2018large,rastogi2017scalable,nouri2018toward}, they may generalize to a new slot by fine-tuning their models but no experiments were done in their works.

\begin{figure}[tp]
  \includegraphics[width=0.48\textwidth]{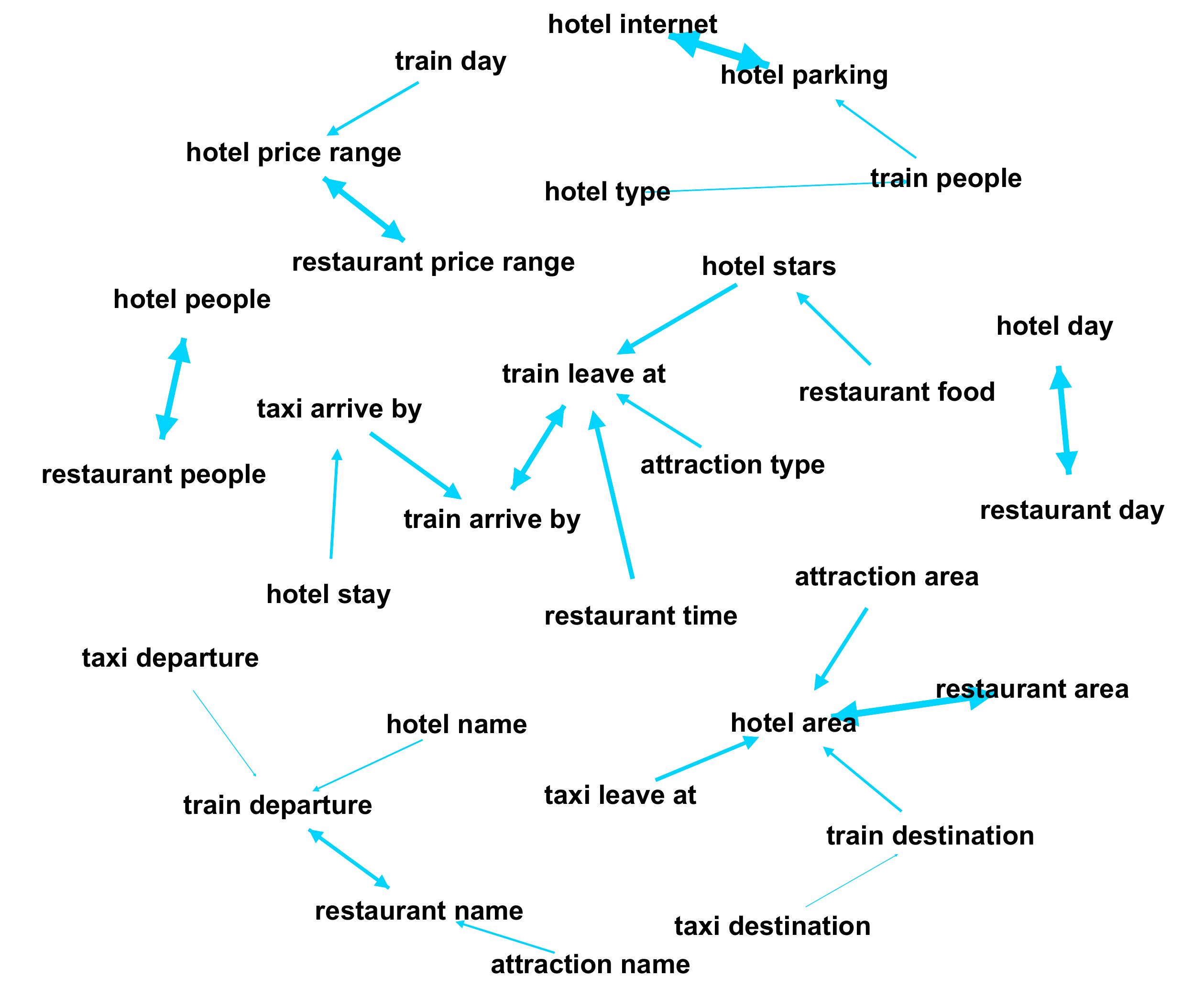}
  \caption{The similarity between slot embeddings, for each edge, the target slot has the highest similarity with source node, and the thickness of the edge indicates the weight and the similarity.
  }
  \label{figure3}
  \end{figure}

In our POGD model, parameters are shared across all slots to share knowledge, when a new slot is added to ontology, with a few data and training epochs, the model could get a satisfactory performance. Furthermore, we find that the similarity between embeddings of slots will train to be consistent with the similarity of the slot values' \textit{entity types}, we verify this by computing the cosine similarity between slots embeddings. We keep the most similar slot for each slot, and use the similarity between their embeddings as weights, then construct a directed graph. We plot the graph using Gephi\footnote{https://gephi.org/} with Fruchterman–Reingold algorithm \cite{fruchterman1991graph}, as shown in Figure~\ref{figure3}. We can find that a pair of slots have a higher similarity when their corresponding values have a more similar entity type. Experiments show this phenomenon can be used to improve the convergence speed of our model.

  \begin{figure}[tp]
    \includegraphics[width=0.48\textwidth]{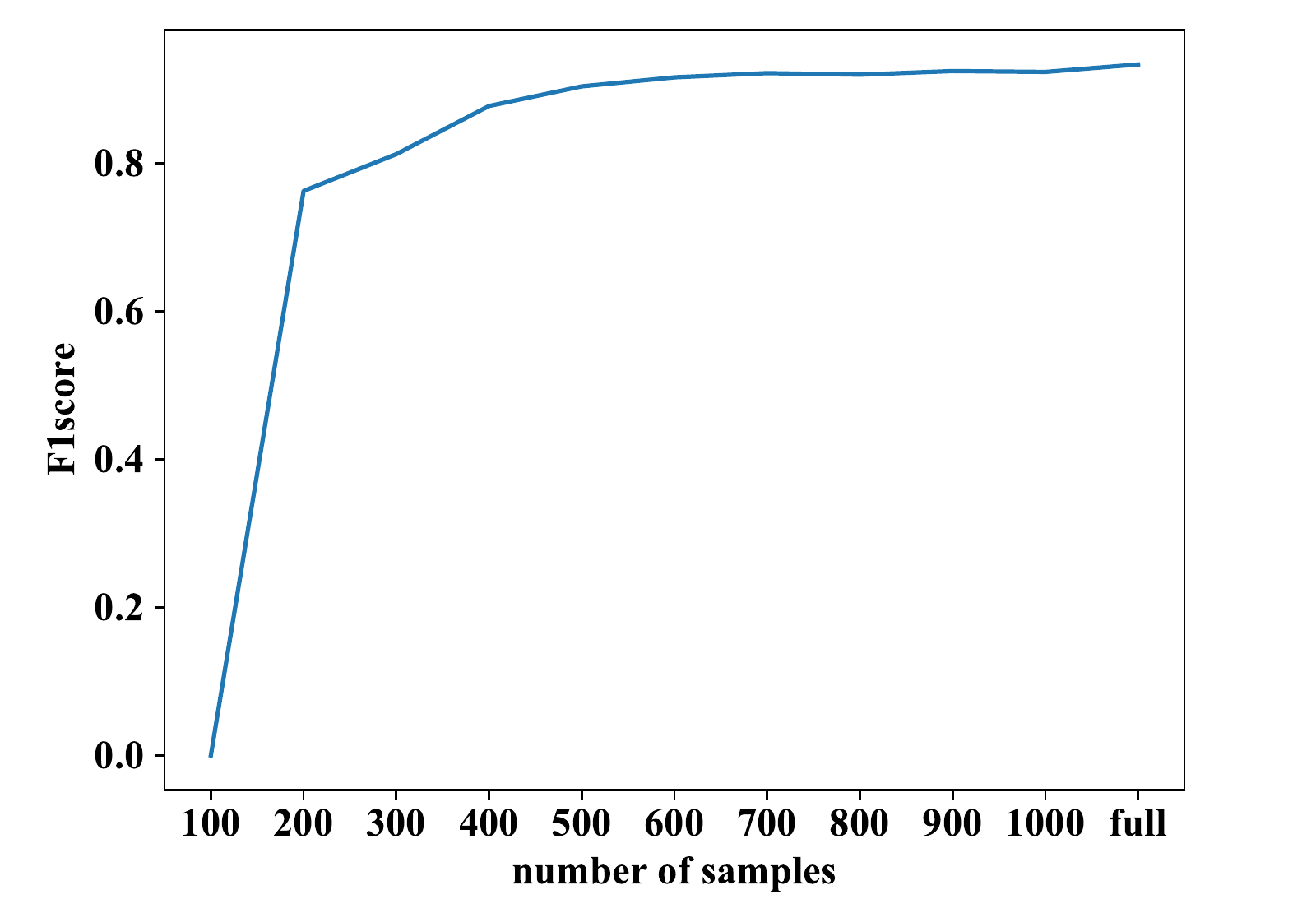}
    \caption{Results of few-shot learning. At each point we train the model for 10 epochs and select the best, \textit{full} at the end of abscissa means we use full data (19651 examples) to train the model.}
    \label{few_shot}
    \end{figure}

    \begin{figure}[!hb]
      \includegraphics[width=0.48\textwidth]{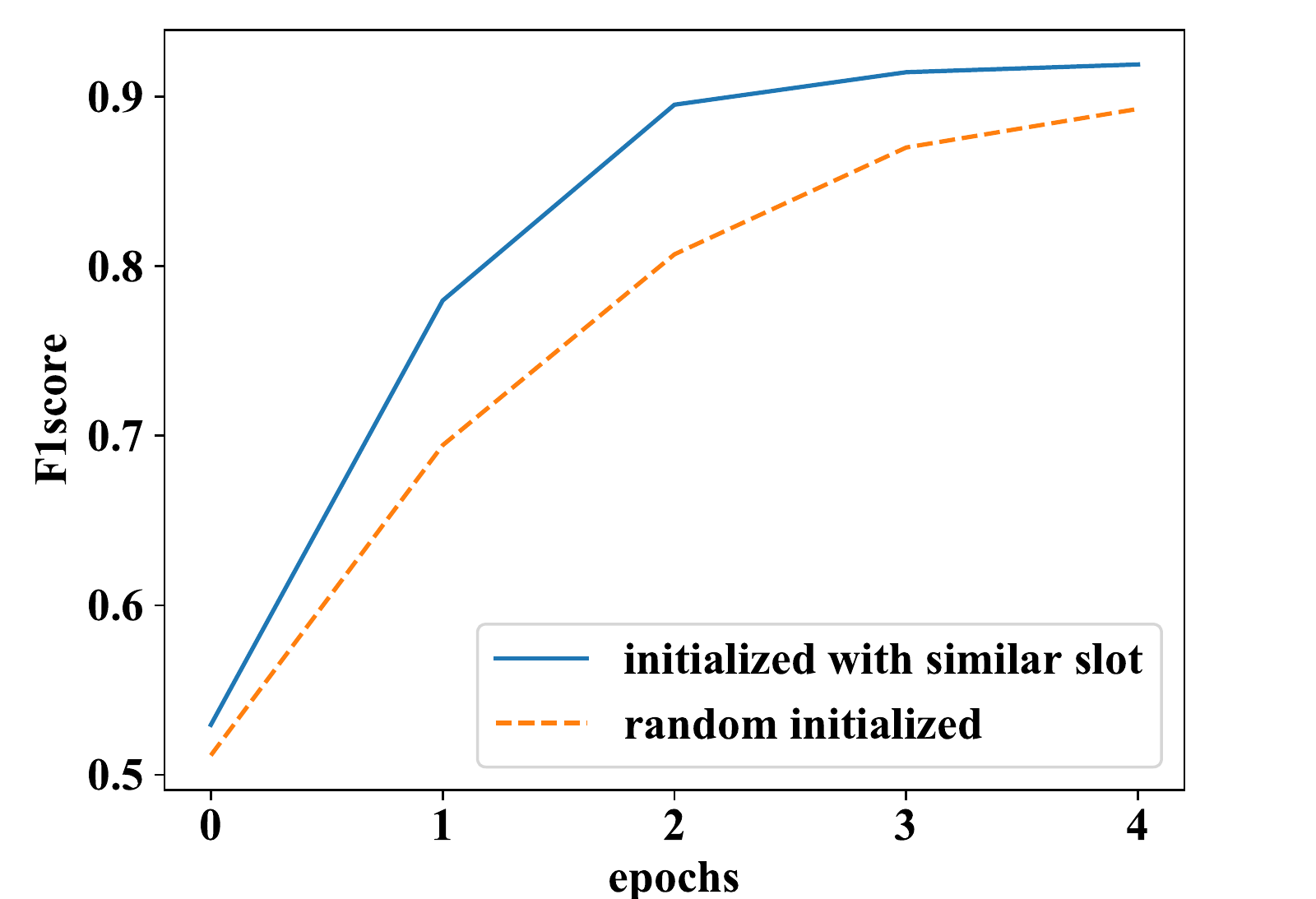}
      \caption{Results of convergence speed tests. The results here are generated by fine-tuning the pretrained model with only 700 of 19651 examples (including negative examples) in 5 epochs. }
      \label{conv_speed}
    \end{figure}

   The following experiments are performed to show the ability of few-shot learning and the convergence speed of POGD. In this subsection, data of 30 slots from 5 domains with 1:10 negative sampling rate are used. We use data of 29 slots except \texttt{train departure} to train a model 10 epochs as a pretrained model, and use \texttt{train departure}'s data to evaluate. We use this slot because it is complex enough and has less overlapping values with its similar slot---\texttt{restaurant name}, which ensures the effectiveness of experiments.

  To show the capability of few-shot learning, we randomly choose training examples from data of slot \texttt{train departure} which contains 19651 examples in total (including negative examples), after fine-tuning the pretrained model using these examples 10 epochs, we choose the best performance and get results shown in Figure~\ref{few_shot}. We find that with only 700 of 19651 (\textbf{3.56\%}) training examples we could get nearly the same results as using full data.

  We compare the convergence speed between the slot \texttt{train departure} is random initialized and initialized with embeddings of \texttt{restaurant name}, the results shown in Figure~\ref{conv_speed} indicate that initializing with similar slot embeddings can significantly increase the convergence speed. But if we only use the new slot's data to fine-tune the pretrained model, after several epochs the pretrained model may be broken, so an efficient way to learn a new slot is using the slot-specific data and full data alternately to fine-tune the pretrained model, while new slot initialized with a similar slot.

\subsection{Ablation study}
\label{ablationstudy}
We perform ablation experiments to analyze the effectiveness of different
components of POGD. The results of these experiments are shown in Table~\ref{tab:ablation}.
\begin{table}[h]
  \centering
  \begin{tabular}{l|l|l}
  \hline
  models & \begin{tabular}[c]{@{}l@{}}joint goal\\ WoZ 2.0\end{tabular} & \begin{tabular}[c]{@{}l@{}}joint goal\\ MultiWoZ 2.0\end{tabular} \\ \hline
  POGD & 0.8870 & 0.3915  \\ 
  POGD+C &0.9411  & 0.5849 \\ 
  POGD+S & 0.8451 & 0.3206 \\ 
  POGD+C+S & 0.9350 & 0.5942 \\ 
  only Pointer & 0.7357 & 0.5615 \\ 
  only Generator & 0.0334 & 0.0297  \\ \hline
  \end{tabular}
  \caption{Results of ablation experiments of POGD. "+C" means we use the labels of Classifier as its outputs, "+S" means the same for Switcher. Only Pointer (Generator) means Switcher always chose the output from Pointer (Generator) and use the Classifier's labels as its outputs.}
  \label{tab:ablation}
  \end{table}

\textbf{Combining Pointer and Generator we can get a strong performance on 
figuring out values given a correct slot.}
When Classifier uses the labels, we can get a 94.11\% joint goal accuracy on WoZ 2.0 and 58.49\% on MultiWoZ 2.0, which outperforms the previous state-of-the-art\cite{sharma2019improving,nouri2018toward} by 4.91\% and 22.91\%. It is because the binary Switcher lets them handle simplified cases: Pointer only learns to point out values from the user's utterance and Generator only learns to infer implicit values.

\textbf{The Switcher choose the best output from Pointer and Generator.} From the results we notice that when only the Switcher uses the generated labels as outputs, the performance of our model is slightly worse, after analyzing the output of the model, we found two reasons. One is the dialogue state doesn't always update as soon as a value is mentioned by the user, but Switcher's labels are generated turn by turn, if a value's first occurrence is missed, it will not be corrected in the followed turns. For example, a user informs the system he wants to find a restaurant that serves Chinese food, but in some data examples the slot-value pair of \texttt{food=chinese} is added to the dialogue state until the booking is completed. In such cases, generated labels cause all of these turn states go wrong, but Switcher and Classifier may decide to output this pair when it is first mentioned, so after turns of wrong state, it may finally get a correct state (\textit{e.g.}, when booking is completed). Another reason is, a pointable value not always produced by Pointer, it may generate by Generator with higher confidence. Because of these two reasons, Switcher produced unexpected results with better performance than the generated labels.
  \begin{table}[ht]
    \centering
    \begin{tabular}{l|l|l}
    \hline
     & \begin{tabular}[c]{@{}l@{}}Acc\\ WoZ 2.0\end{tabular} & \begin{tabular}[c]{@{}l@{}}Acc\\ MultiWoZ 2.0\end{tabular} \\ \hline
    Switcher & 0.9782 & 0.9837 \\ 
    Classifier & 0.9793  & 0.9888 \\ \hline
    \end{tabular}
    \caption{Performance of Switcher and Classifier.}
    \label{tab:perfsc}
    \end{table}
  
\textbf{The Classifier works well but suffers a lot from error accumulation.}
Even though it has high accuracy as illustrated in Table~\ref{tab:perfsc}, but with the increasing of slots, each percent of classifier errors cause more loss on the joint goal accuracy, because at each turn we iterate all slot and try to figure out their values, but most of them are not relevant to the utterance. This is the biggest problem with our model. We should design it in a more stable way in the future.

\section{Conclusion}
\label{conclusion}
We propose the Point-Or-Generate Dialogue State Tracker (POGD)---a scalable multi-domain dialogue state tracker. POGD can handle unseen values and reduce the effort when new slots added to the ontology. Our model obtains 88.7\% joint goal accuracy on the WoZ 2.0 dataset, and on the large-scale multi-domain dataset MultiWoZ 2.0, it obtains 39.15\% joint goal accuracy, outperforming prior work by 3.57\%. 
\bibliography{emnlp-ijcnlp-2019}
\bibliographystyle{acl_natbib}

\newpage
\appendix

\section{Appendix}

\subsection{Data generation}
POGD is designed to predict the turn goals, but in MultiWoZ 2.0, it only contains the full dialogue states, so we need to generate turn goals labels. The turn goals labels are generated as follows, at the first turn of dialogue the state is equal to the turn goals, for the following turns, we simply delete the slot-value pairs that appeared in previous turns. If a value doesn't occur in the dialogue state on its first occurrence, we will add it to the turn state where it was mentioned. This operation only performed in WoZ 2.0, in MultiWoZ 2.0, a value may use for several slots, this operation may lead to further errors. One important thing, no modification was performed on the test set.

\subsection{Parameters}

Words are represented using 300 dimensional embeddings, and each slot is mapped to a randomly initialized 100 dimensional embedding.  \textbf{LSTMs:} all LSTMs in POGD are one layer BiLSTM and their hidden state size 1s 128. \textbf{Linears:} $Linear^{P_{1}}$, $Linear^{P_{2}}$ and $Linear^{G}$ have the same output size of 2 * BiLSTMs' hidden size to perform attention. The output size of $Linear^{S}$ and $Linear^{C}$ is equal to their input dimensions. \textbf{MLPs}: $MLP^{S}$ and $MLP^{C}$ have the same architecture as described in Section 3, they have hidden layer sizes the same as LSTMs' hidden size. $MLP^{G}$ is a $\mathrm{MLP^{2}_{ReLU}}$ and has a hidden size of 2 * dimension of words embeddings.

During the training process on WoZ 2.0 dataset, due to such little data, instead of $\mathrm{MLP^{3}_{ReLU}}$ on MultiWoZ 2.0 dataset, Classifier and Switcher are $\mathrm{MLP^{2}_{ReLU}}$ and the $MLP^{G}$ is not used. We set the hidden state of BiLSTMs to 150 to perform $Attn^{G}_{V}$ without $MLP^{G}$, because 2 * LSTM's hidden size should equals to the embeddings of values. 

\subsection{Tricks}

During the training process, we used a dynamic batch size to adjust the learning rate. We changed the batch size every 30 epochs on WoZ 2.0 and 5 epochs on MultiWoZ, and each time batch size changed, we will load the best model parameters up to now and continue to training. We found that this trick accelerated the convergence and slightly increased the performance. 

\end{document}